\documentclass{article}
\usepackage[nonatbib,preprint]{arxiv}
\usepackage[utf8]{inputenc} %
\usepackage[T1]{fontenc}    %
\usepackage[colorlinks,citecolor=blue]{hyperref}       %
\usepackage{url}            %
\usepackage{booktabs}       %
\usepackage{amsfonts}       %
\usepackage{nicefrac}       %
\usepackage{microtype}      %
\usepackage{xcolor}         %
\usepackage{subcaption}
\usepackage{multirow} %
\usepackage{caption} %
\usepackage{hhline} %
\usepackage{tikz}
\usepackage{tabularx} 
\usepackage{amssymb} 
\usetikzlibrary{calc}
\usepackage{arydshln}
\usepackage{enumitem}
\setitemize{noitemsep,topsep=0.pt,parsep=0pt,partopsep=0pt}

\usepackage{amsmath,amssymb,amsfonts,bm,mathtools}
\usepackage{amsthm}
\usepackage{graphicx}
\usepackage{flexisym}

\theoremstyle{plain}

\theoremstyle{definition}

\theoremstyle{remark}

\usepackage[ruled,linesnumbered]{algorithm2e}
\SetKwComment{Comment}{/* }{ */}
\usepackage{wasysym}
\usepackage{mathtools}
\usepackage{authblk}
\usepackage{algorithmic}
\usepackage{subcaption}
\usepackage{wrapfig}
\usepackage{resizegather}

\def\eqref#1{equation~\ref{#1}}

\numberwithin{equation}{section}

\def\1{\bm{1}}

\DeclareMathAlphabet{\mathsfit}{\encodingdefault}{\sfdefault}{m}{sl}
\SetMathAlphabet{\mathsfit}{bold}{\encodingdefault}{\sfdefault}{bx}{n}


\title{AODDiff: Probabilistic Reconstruction of Aerosol Optical Depth via Diffusion-based Bayesian Inference }

\author{%
  Linhao Fan$^{1}$~~~~Hongqiang Fang$^2$~~~~Jingyang Dai$^1$~~~~Yong Jiang$^1$~~~~Qixing Zhang$^1$\thanks{Corresponding author}\\ 
  
  \vspace{-0.5em} 
  
  \small 
  
  $^1$State Key Laboratory of Fire Science, University of Science and Technology of China \\
  $^2$Fire Research Division, National Institute of Standards and Technology
  
  \texttt{\{fan7tree,djy2001\}@mail.ustc.edu.cn}\\
  \texttt{hongqiang.fang@nist.gov}\\
  \texttt{\{yjjiang, qixing\}@ustc.edu.cn}
}

\usepackage[nonatbib,preprint]{arxiv} 

\usepackage[numbers,sort&compress]{natbib}

\begin{document}
\maketitle

\begin{abstract}
High-quality reconstruction of Aerosol Optical Depth (AOD) fields is critical for Atmosphere monitoring, yet current models remain constrained by the scarcity of complete training data and a lack of uncertainty quantification.To address these limitations, we propose AODDiff, a probabilistic reconstruction framework based on diffusion-based Bayesian inference. By leveraging the learned spatiotemporal probability distribution of the AOD field as a generative prior, this framework can be flexibly adapted to various reconstruction tasks without requiring task-specific retraining. We first introduce a corruption-aware training strategy to learns a spatiotemporal AOD prior solely from naturally incomplete data. Subsequently, we employ a decoupled annealing posterior sampling strategy that enables the more effective and integration of heterogeneous observations as constraints to guide the generation process. We validate the proposed framework through extensive experiments on Reanalysis data. Results across downscaling and inpainting tasks confirm the efficacy and robustness of AODDiff, specifically demonstrating its advantage in maintaining high spatial spectral fidelity. Furthermore, as a generative model, AODDiff inherently enables uncertainty quantification via multiple sampling, offering critical confidence metrics for downstream applications.

\end{abstract}

\section{Introduction}

Aerosols are vital atmospheric components that profoundly influence surface radiation balance and regional climate through the scattering and absorption of solar radiation \cite{ramanathan2001aerosols,prather2008analysis,boucher2015atmospheric}. Among aerosol characterization parameters, Aerosol Optical Depth (AOD) \cite{holben1998aeronet} serves as a fundamental and widely adopted metric. It quantifies the total extinction effect of aerosols along the atmospheric column and serves as a fundamental parameter in atmospheric physics and climate modeling \cite{kaufman2002satellite,ramanathan2001aerosols}. This importance becomes even more pronounced during sudden high-intensity aerosol events, such as wildfire smoke outbreaks, where high spatiotemporal resolution AOD observations are crucial.

However, AOD retrieval relies on remote sensing satellites and sparse ground networks, which struggle to achieve high resolution simultaneously in both spatial and temporal dimensions \cite{wei2020satellite}. For instance, polar-orbiting satellites (e.g., MODIS and VIIRS) offer kilometer-level spatial resolution but represent limited temporal coverage (1 or 2 passes daily) \cite{levy2013collection}. Conversely, geostationary satellites (e.g., Himawari-8) provide high temporal resolution (15 minutes) but coarser spatial resolution and limited coverage \cite{bessho2016introduction}. Moreover, the efficacy of AOD retrieval is frequently impeded by physical constraints, including cloud contamination and heavy aerosol loading (e.g., dense wildfire smoke plumes \cite{li2025using}), resulting in pervasive and systematic missing values in observations. These systematic spatiotemporal gaps severely restrict downstream applications, such as fire emission estimation \cite{ye2023impact,li2025using} and air quality forecasting \cite{lee2022air}. Consequently, efficient AOD field reconstruction is critical. Mathematically, this represents an challenging inverse problem: recovering a high-dimensional, complete state $x$ from incomplete, noisy observations $y$. This problem domain typically addresses three interrelated challenges: missing data imputation, spatial downscaling, and temporal interpolation \cite{eyring2024pushing}.

To tackle these complex tasks, several distinct paradigms have emerged, among which Data Assimilation (DA) systems have long served as the standard approach for this inverse problem in atmospheric science. Current state-of-the-art operational aerosol reanalysis products, such as NASA’s MERRA-2 \cite{gelaro2017modern} and ECMWF’s CAMS \cite{peuch2022copernicus}, primarily utilize three- or four-dimensional variational assimilation techniques to integrate discrete observationas
into numerical models governed by fluid dynamics and atmospheric chemistry equations. While theoretically rigorous, DA systems face prohibitive computational costs \cite{maulik2022efficient}, necessitating significant compromises in spatial resolution. To address these computational bottlenecks, researchers have transitioned toward more flexible statistical learning paradigms. Early efforts in AOD gap-filling relied on geostatistical interpolation methods \cite{zhang2022gap,yang2018filling}, such as Inverse Distance Weighting (IDW) and Kriging. These approaches, typically predicated on assumptions of spatial stationarity, often struggle with real-world aerosol fields characterized by pronounced non-stationarity and anisotropy. Consequently, those interpolation methods act effectively as low-pass filters: they tend to blur the high-frequency textures and fine-scale details essential for accurate representation \cite{chen2022novel}.

Recently, deep learning (DL) has been widely adopted for AOD reconstruction due to its robust nonlinear mapping capabilities, achieving accuracies that significantly surpass traditional methods \cite{bai2024lghap,wei2021multi,liang2023reconstructing}. However, current DL models, which treat AOD reconstruction as a standard regression task, face some inherent deficiencies when addressing this inverse problem. First, the supervised learning paradigm suffers from data dependencies and limited task generalizability. The foundation of these models is the availability of high-quality paired datasets. The inherent systematic missingness of AOD observations makes constructing such enough paired datasets extremely difficult. Furthermore, this strict reliance on fixed training pairs confines models to a single optimization objective. Consequently, they often lack the generalizability to handle unified applications across diverse tasks, such as spatiotemporal imputation, interpolation, and multi-source fusion. Second, at the methodological level, existing frameworks adopt a deterministic, point-estimate learning paradigm. From a statistical standpoint, such formulations implicitly encourage the model to approximate the central tendency of the conditional posterior distribution. Since a single degraded input corresponds to an infinite manifold of plausible high-frequency solutions, this paradigm inevitably yields predictions that resemble an "average" over all possibilities \cite{wang2022review}. As a consequence, critical high-frequency textures and extreme pollution peaks are systematically suppressed, manifesting as a pronounced attenuation of the Power Spectral Density (PSD) in high-wavenumber regions \cite{bischoff2024unpaired}. Third, deterministic models yield only point estimates, neglecting the quantification of uncertainty. This limitation is critical for risk-related downstream tasks (e.g., wildfire monitoring), where the lack of predictive confidence intervals can lead to an underestimation of potential hazards.

To address these limitations, this study reframes AOD reconstruction from a standard regression task to a conditional probabilistic generation task. Leveraging the rapid development of diffusion models \cite{yang2023diffusion, lai2025principles}, this approach provides a powerful paradigm for learning the probability distribution of high-dimensional data. It naturally integrates observational constraints during the sampling process, enabling direct sampling from the posterior distribution \cite{chung2022diffusion}. This diffusion-based Bayesian inverse problem-solving framework \cite{daras2024survey} has achieved state-of-the-art results in  image \cite{luo2025taming}, medical imaging \cite{kazerouni2023diffusion} and flow field \cite{li2024learning} reconstruction and is progressively being introduced into meteorological observation reconstruction \cite{tu2025satellite,schmidt2025generative,hess2025fast,mardani2025residual,ling2024diffusion}. This study applies this framework to the AOD reconstruction task, developing a probabilistic generative reconstruction model tailored to the characteristics of AOD data. The main content and contributions are summarized as follows:

\begin{itemize}
        \item We propose AODDiff, a diffusion-based Bayesian inference framework that learns the spatiotemporal AOD distribution as a generative prior, allowing a unified and retraining-free solution to diverse AOD reconstruction tasks.
        \item We introduce a corruption-aware training strategy designed to learn a spatiotemporal AOD prior directly from incomplete observations.
        \item We employ a decoupled annealing posterior sampling strategy that enables the more effective integration of heterogeneous observations as constraints to guide the generation process.
\end{itemize}

\section{Problem Setting and Data Construction}\label{sec2}
\subsection{Formulation of AOD Reconstruction Problem}\label{subsec2_1}

We model the AOD as a continuous spatiotemporal field, denoted by $\mathbf{X}$. The observed data, $\mathbf{Y}$, is derived from $\mathbf{X}$ through a physical degradation process, formally defined as:

$$\mathbf{Y} = \mathcal{A}(\mathbf{X}) + \epsilon, \quad \epsilon \sim \mathcal{N}(0, \sigma_y^2 \mathbf{I})$$

where $\mathcal{A}$ is an observation operator representing physical degradation (e.g., downsampling or masking). The term $\epsilon$ represents observational noise, modeled as Gaussian white noise with zero mean and variance $\sigma_y^2$. The AOD field reconstruction problem is therefore framed as an inverse problem: given the observations $\mathbf{Y}$ and the operator $\mathcal{A}$, the objective is to estimate the complete AOD field $\mathbf{X}$ at the target spatiotemporal resolution.

However, evaluating the performance of reconstruction models in real-world scenarios faces a fundamental challenge: a "ground truth" AOD field that possesses both the ideal spatiotemporal resolution and is entirely gap-free is practically unobtainable. To enable quantitative and controlled evaluation, this study thus formulates a proxy task where the ground truth field is explicitly available.

To implement this proxy task, it is essential to use a dataset that serves as a reliable surrogate for the true atmospheric state. Specifically, we adopt reanalysis data as the reference data, selected for its spatiotemporal continuity and physical consistency. It facilitates for the simulation of observational degradation processes with high precision and control. Critically, when the observation operators ($\mathcal{A}$) are defined to closely reflect actual sensing constraints, the resulting evaluation becomes directly informative for practical reconstruction tasks. Successful reconstruction, where the estimated field $\hat{\mathbf{X}}$ accurately matches the reference data $\mathbf{X}_{GT}$ in this high-fidelity proxy environment, demonstrates the model's potential to handle real-world tasks. This setup effectively transforms the problem into a controllable,quantitatively evaluable framework, thereby guaranteeing the reliability and practical relevance of the model validation and subsequent conclusions.The specific dataset used for this study will be detailed in Section \ref{subsec2_2}, while the precise definition of the observation operator $\mathcal{A}$ is provided in Section \ref{subsec2_3}.

\subsection{Training Data Construction}\label{subsec2_2}

\begin{figure}[ht]
  \centering
  \includegraphics[width=0.75\textwidth]{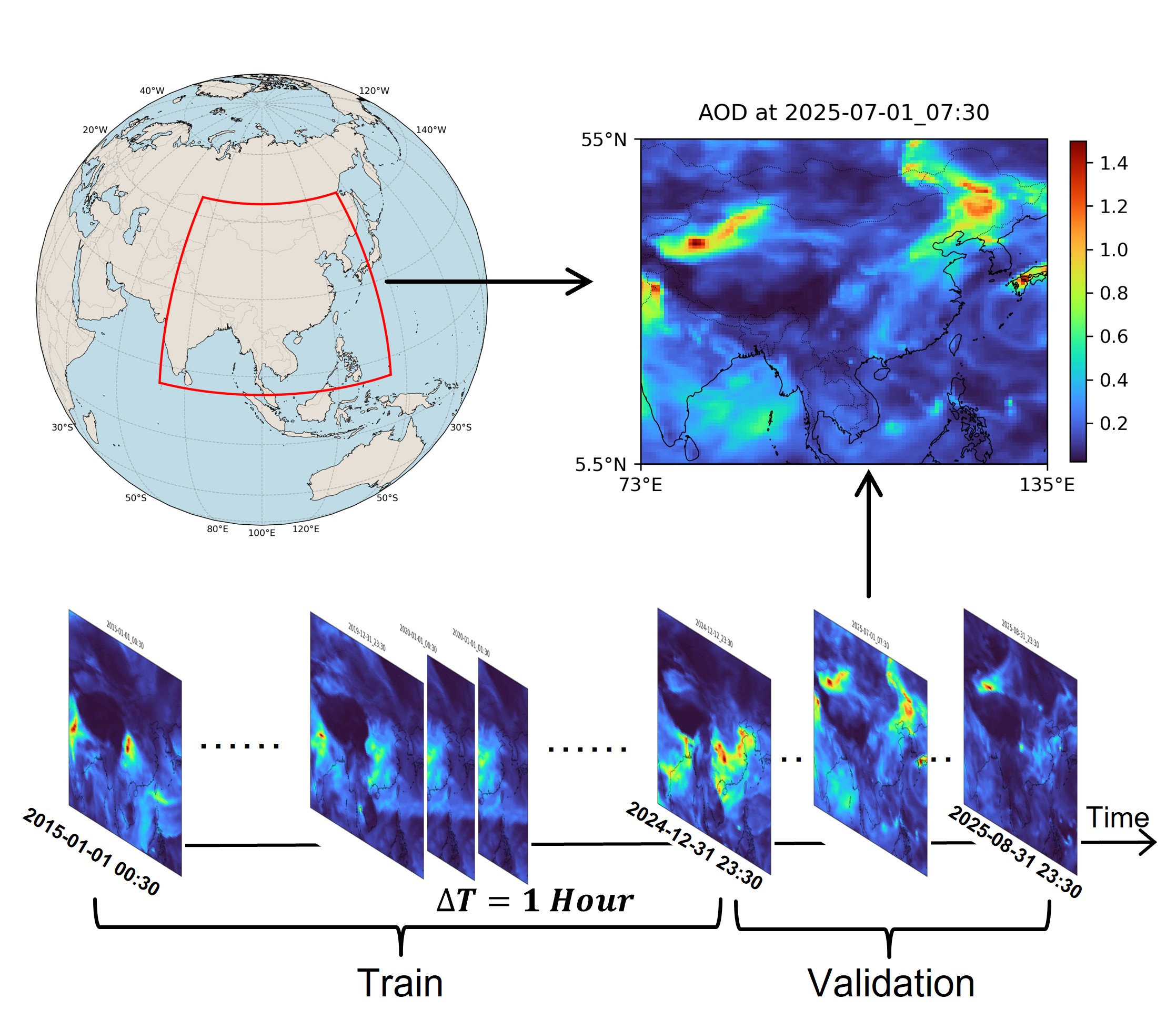}
  \caption{Spatiotemporal Scope and Train/Validation Split of the AOD Dataset}
  \label{fig2_1}
\end{figure}

In this study, we employ the MERRA-2 as the reference data. MERRA-2 is a state-of-the-art reanalysis product and one of the most widely adopted datasets in atmospheric science and aerosol research, offering spatiotemporally continuous and physically consistent fields. Specifically, the AOD data are obtained from the Total Aerosol Optical Depth at 550 nm within the hourly time-averaged aerosol collection. This dataset features a horizontal resolution of $0.5^\circ$ in latitude and $0.625^\circ$ in longitude, with an hourly temporal resolution. The study domain covers East and South Asia ($5.5^\circ\text{N}$–$55^\circ\text{N}$, $73^\circ\text{E}$–$135^\circ\text{E}$), as illustrated in Figure \ref{fig2_1}. This region exhibits substantial spatiotemporal variability in AOD, making it a suitable testbed for evaluating reconstruction algorithms. For model training, we compile a 10-year dataset spanning 2015–2024, while data from 2025 is reserved exclusively for model validation.

To effectively handle the heavy-tailed distribution and extreme positive outliers inherent in the AOD data, we implement a robust standardization strategy. The logarithmic transformation is applied to the raw AOD data to suppress the effects of outliers and manage the heavy tailed distribution, followed by a quantile-based robust scaling:

$$\mathbf{X}_{\text{log}} = \log(1 + \mathbf{X})$$
$$\mathbf{X}_{\text{norm}} = 2 \times \left( \frac{\mathbf{X}_{\text{log}} - Q_{0.01}}{Q_{0.99} - Q_{0.01}} \right) - 1$$

Here, $Q_{0.01}$ and $Q_{0.99}$ denote the $1^{\text{st}}$ and $99^{\text{th}}$ percentile values of the log-transformed AOD field, estimated exclusively from the entire training dataset. This robust scaling ensures that the vast majority of the data is mapped to the target range, thereby achieving zero-centering and maintaining resilience against extreme outliers.

\subsection{Observation Operators}\label{subsec2_3}

\begin{figure}[ht]
  \centering
  \includegraphics[width=0.8\textwidth]{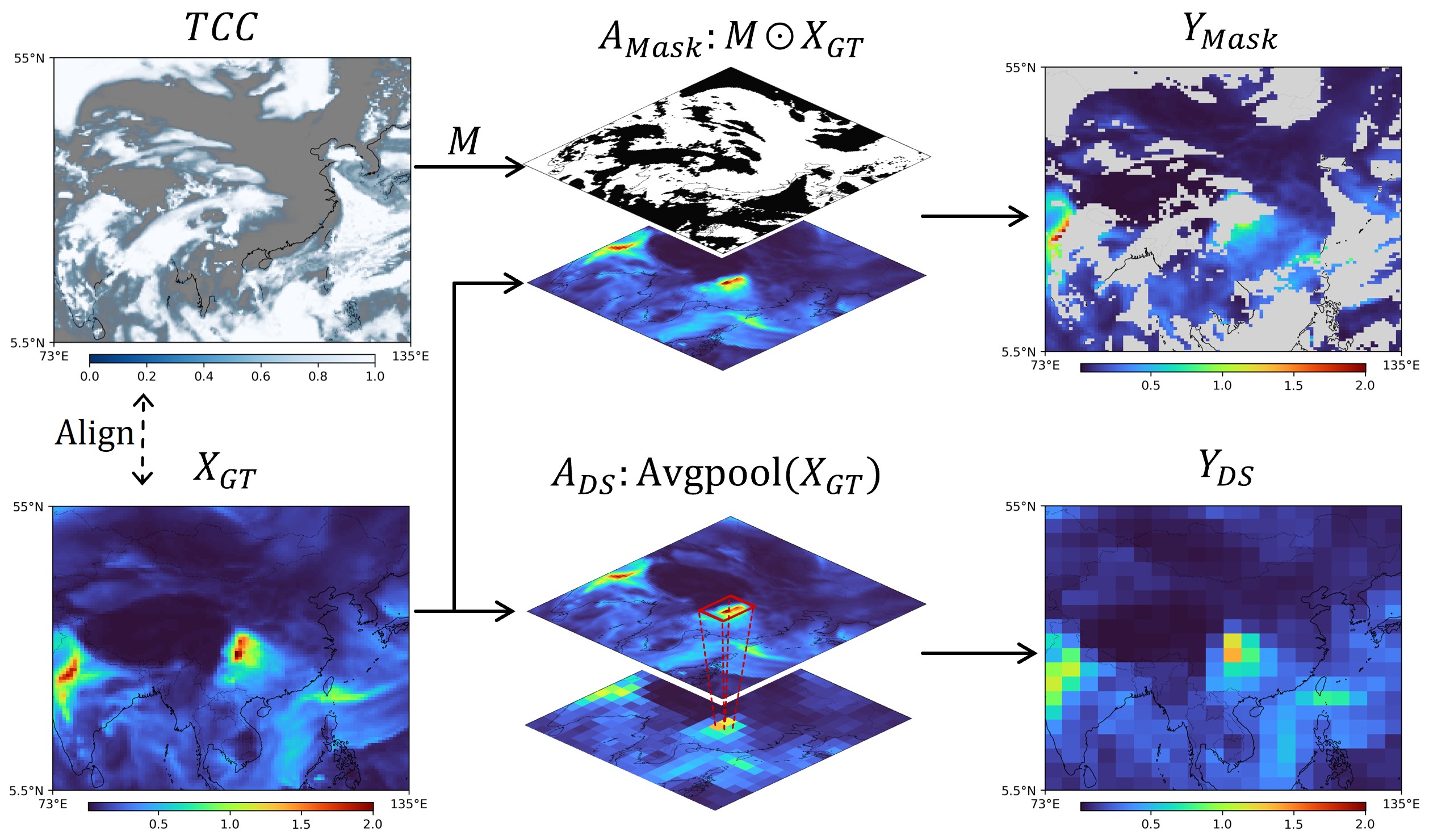}
  \caption{Workflow of the observation operators. The reference field $\mathbf{X}_{\text{GT}}$ is processed by the masking operator $\mathcal{A}_{Mask}$ to simulate cloud-occluded gaps, and the downsampling operator $\mathcal{A}_{DS}$ to generate low-resolution representations.}
  \label{fig2}
\end{figure}

This study defines two corresponding observation operators, $\mathcal{A}_{Mask}$ and $\mathcal{A}_{DS}$, to specifically tackle the primary AOD reconstruction tasks of missing data imputation and downscaling, respectively. These operators are applied to the reference data $\mathbf{X}_{\text{GT}}$ to generate simulated observations that mimic realistic degradation processes, such as cloud-induced gaps and spatial coarsening, as illustrated in the workflow of Figure \ref{fig2}.

\subsubsection{Masking Operator}

Missing AOD data in real satellite products are dominated by cloud obstruction, producing complex spatial structures and strong temporal correlations that simple random or block masks cannot replicate. To generate physically realistic missingness patterns $\mathbf{M}$, we derive them from the Total Cloud Cover (TCC) variable of the ERA5 reanalysis dataset \cite{hersbach2020era5}. Hourly ERA5 TCC fields (2015–2024) are first resampled to the exact grid of $\mathbf{X}_{GT}$, yielding $\text{TCC}(t,i,j)$ for each time step $t$ and grid cell $(i,j)$. Here, $i$ and $j$ denote the indices for latitude and longitude, respectively. A cloud-threshold parameter $\gamma\in[0,1]$ is then applied: grid cells with mean cloud cover exceeding $\gamma$ are marked as missing, while those below the threshold are retained:

$$\mathbf{M}(t, i, j) = \begin{cases} 0 & \text{if } \text{TCC}(t, i, j) > \gamma\\ 1 & \text{otherwise}\end{cases}$$

Because this cloud-based mask preserves the true spatial morphology and temporal evolution of real missing AOD, it provides a high-fidelity proxy for evaluating inpainting performance.The final masked observation is obtained via element-wise degradation:

$$\mathbf{Y}_{\text{Mask}} = \mathcal{A}_{\text{Mask}}(\mathbf{X}_{GT})= \mathbf{M} \odot \mathbf{X}_{GT} +\epsilon_m$$

where $\epsilon_m \sim \mathcal{N}(0, \sigma_m^2)$ represents the independent additive Gaussian white noise that simulates sensor measurement errors in the masking process.

\subsubsection{Downsampling Operator}

Low-resolution satellite retrievals typically represent aggregated signals over a sensor’s instantaneous field of view. To emulate this physical process, spatial downsampling is implemented using average pooling. Let $s_{\text{step}}$ denote the spatial downscaling factor. A low-resolution field is produced by applying an $s_{\text{step}}\times s_{\text{step}}$ average pooling operation to the high-resolution ground truth. If temporal sampling is also reduced, a temporal step factor $t_{\text{step}}$ is used to subsample the pooled sequence:

$$\mathbf{Y}_{DS} = \mathcal{A}_{DS}(\mathbf{X}_{GT})  = \text{AvgPool}_{s_{step}}(\mathbf{X}_{GT})_{t \in \{t_{step}\}} + \epsilon_m$$

where $\mathbf{Y}_{DS}$ denotes the simulated low-resolution observation. The operator performs spatial average pooling with factor $s_{step}$ and temporal subsampling at every $t_{step}$ interval.

\section{Method}\label{sec:method}

\subsection{Diffusion-based Bayesian Inference Framework}\label{subsec3_1}
In this study, the reconstruction of the AOD field is formulated as a probabilistic generative problem, specifically as solving and sampling from the posterior distribution $p(X\mid Y)$, where $x$ denotes the target AOD field and $y$ represents the available observational data.

To model the probability distribution of the high-dimensional AOD field, we leverage a framework based on Score-Based Diffusion Model \cite{song2020score}. It is built upon a pair of interconnected processes, which govern how the model learns to characterize a complex distribution.

\begin{enumerate}
    \item Forward Diffusion Process: A Stochastic Differential Equation (SDE) is used to progressively add Gaussian noise to the data $\mathbf{z}_0 \sim p(\mathbf{z})$, driving it into a standard normal distribution $\mathbf{z}_1 \sim \mathcal{N}(\mathbf{0},\mathbf{I})$:
    
    $$\mathrm{d}\mathbf{z}_{\tau} = \mathbf{f}(\mathbf{z}_{\tau},\tau)\,\mathrm{d}\tau + g(\tau)\,\mathrm{d}\mathbf{w}_{\tau}$$

    Here, $\mathbf{w}_{\tau}$ is the standard Wiener process, and $\mathbf{f}$ and $g$ are the predefined drift and diffusion coefficients. And $\tau$ denotes the continuous diffusion time, which represents the progress of the diffusion process.
    
    \item Reverse Sampling Process:The process of recovering the data $\mathbf{z}_0$ from the noise $\mathbf{z}_1$ is governed by the time-reversed SDE:
    
    $$\mathrm{d}\mathbf{z}_{\tau} = \bigl[\mathbf{f}(\mathbf{z}_{\tau},\tau)-g(\tau)^2\nabla_{\mathbf{z}_{\tau}}\log p(\mathbf{z}_{\tau})\bigr]\,\mathrm{d}\tau + g(\tau)\,\mathrm{d}\bar{\mathbf{w}}_{\tau}$$

    The crucial term is the Score Function $\nabla_{\mathbf{z}_{\tau}}\log p_{\tau}(\mathbf{z}_{\tau})$, the gradient of the log probability density. It provides the necessary gradient information to effectively reverse the process. A parameterized neural network $\mathbf{s}_\theta(\mathbf{z},\tau)$ is trained to approximate this function, thereby learning the structure of the target distribution $p(\mathbf{z})$
\end{enumerate}

Considering that directly modeling the conditional distribution $p(\mathbf{x}\mid \mathbf{y})$ is computationally challenging, we further adopt the diffusion-based Bayesian inverse problem-solving framework \cite{chung2025diffusion}. It decomposes the posterior score into two components:

$$\nabla_{\mathbf{x}_{\tau}}\log p(\mathbf{x}_{\tau} \mid \mathbf{y}) = \underbrace{\nabla_{\mathbf{x}_{\tau}}\log p(\mathbf{x}_{\tau})}_{\text{Prior Score}} + \underbrace{\nabla_{\mathbf{x}_{\tau}}\log p(\mathbf{y}\mid \mathbf{x}_{\tau})}_{\text{Likelihood Score}}$$

The Prior Score is provided by the trained diffusion model (the neural network $\mathbf{s}_\theta$), which encodes the prior knowledge of the AOD field. The Likelihood Score is the gradient of the likelihood function with respect to the measurement data, serving to inject the observational constraints $\mathbf{y}$ into the sampling process. And this framework offers high flexibility: the Likelihood Score is independent of the specific observation operator, allowing the method to naturally accommodate diverse types of observations without retraining the core prior model.

This decomposition strategy allows us to sequentially address the modeling challenges: Section \ref{subsec3_2} details the unconditional prior score modeling within the AODDiff framework, and Section \ref{subsec3_3} elucidates the mechanism for providing effective posterior guidance during the sampling phase.

\subsection{Learning AOD Prior Distributions through Score-based Diffusion Model}\label{subsec3_2}

Considering the inherent incompleteness of AOD data, this study dispenses with the assumption of complete data availability for model training. Instead, the prior distribution of AOD is learned solely from the available data. And the AOD field behaves as a dynamic system with complex spatiotemporal couplings. To capture temporal dependencies without the intractability of modeling infinite sequences, we adopt the Pseudo-Markov Blanket assumption \cite{rozet2023score}. This assumption posits that the state $\mathbf{x}_i$ at any time $i$ is primarily dependent on the states within a finite, local time window. Consequently, we define our modeling unit as the data cube $\mathbf{x}_{i-k:i+k}$ spanning $w=2k+1$ time steps. The network $\mathbf{s}_\theta$ is then tasked with approximating the joint Score Function of this local spatiotemporal segment.

\begin{figure}[ht]
  \centering
  \includegraphics[width=0.9\textwidth]{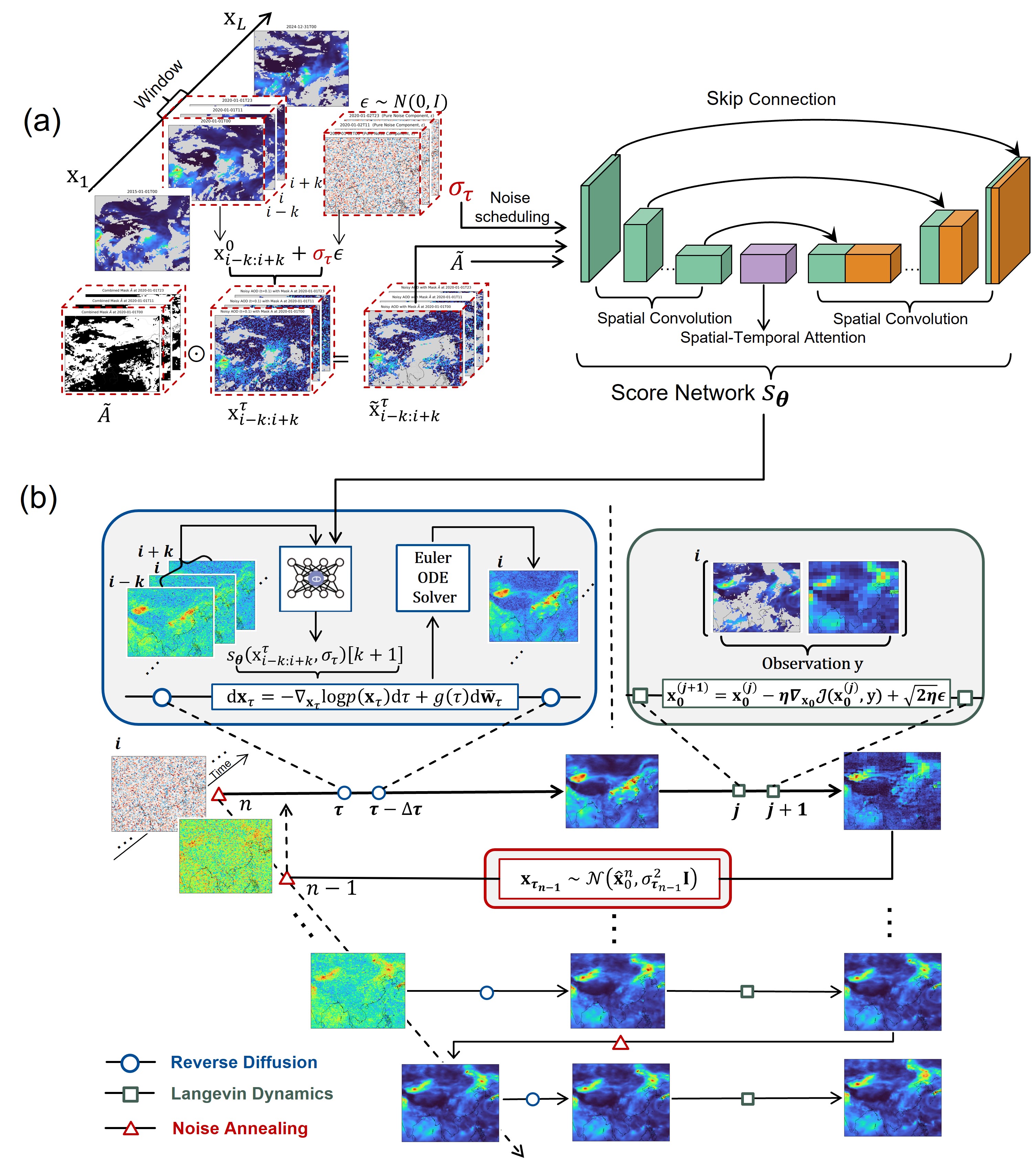}
  \caption{Overall framework of the AODDiff: (a) The corruption-aware training process, where the score network learns the spatiotemporal prior from incomplete observations. (b) The inference process, which employs decoupled annealing posterior sampling strategy to reconstruct the AOD field.}
  \label{fig3}
\end{figure}

To handle missing values during training, we introduce a corruption-aware training scheme inspired by the Ambient Diffusion framework \cite{daras2023ambient}, visually detailed in Figure \ref{fig3}(a).

Given a data segment $\mathbf{x}^0_{i-k:i+k}$ (containing original observational gaps), we first sample a noise level $\tau \sim P_{\text{EDM}}(\tau)$ and add Gaussian noise $\boldsymbol{\epsilon}$ according to the EDM \cite{karras2022elucidating} formulation, yielding the noisy state $\mathbf{x}^\tau_{i-k:i+k}$. We define $A$ as the inherent observation mask (where $1$ denotes observed and $0$ denotes missing). Simultaneously, we generate an additional random dropout mask $B$. These are combined to form the training mask $\tilde{A} = A \odot B$ (logical intersection of valid regions). The final input to the network is the noisy data masked by $\tilde{A}$, denoted as $\tilde{\mathbf{x}}^\tau_{i-k:i+k} = \mathbf{x}^\tau_{i-k:i+k} \odot \tilde{A}$. Crucially, the mask $\tilde{A}$ is concatenated with the data $\tilde{\mathbf{x}}^\tau_{i-k:i+k}$ and fed into the network. This concatenation serves to explicitly provide the model with the location of missing regions (the corruption operator), which is essential for the network to perform conditional score estimation.

The Local Score Network $\mathbf{s}_\theta$, illustrated in Figure \ref{fig3}(a), employs a U-Net architecture adapted for spatiotemporal data \cite{ho2022imagen,li2024learning}. It treats the time window $w$ as a depth dimension, processing the input effectively like a 3D volume. The encoder progressively reduces the spatial resolution to extract multi-scale features, while the decoder reconstructs the signal. Skip connections preserve high-frequency details. To effectively model long-range dependencies within the window, we integrate a Spatial-Temporal Attention mechanism at the bottleneck layer, allowing the model to attend to relevant features across both space and time frames.

The network is trained to predict the clean signal $\mathbf{x}^0$ from the corrupted, noisy input. The loss function is a weighted Mean Squared Error, calculated on the regions that were originally observed ($A$). We formulate the Ambient Score Matching loss as:

$$\mathcal{L} = \mathbb{E}_{\tau, \mathbf{x}, \boldsymbol{\epsilon}} \left[ \lambda(\tau) \left\| A \odot \left( \mathbf{s}_\theta(\tilde{\mathbf{x}}^\tau_{i-k:i+k}, \tilde{A}, \tau) - \mathbf{x}^0_{i-k:i+k} \right) \right\|_2^2 \right]$$

where $\lambda(\tau)$ is the weighting function dependent on the noise level. This objective ensures the model learns to hallucinate plausible structures in missing regions consistent with the observed spatiotemporal context.

\subsection{Posterior Sampling Strategy for AOD Conditional Reconstruction}\label{subsec3_3}

Based on the Bayesian Inference framework established in Section \ref{subsec3_1}, our goal is to sample from the posterior distribution $p(\mathbf{x} \mid \mathbf{y})$. To achieve a robust and flexible solution,we adopt the Decoupled Annealing Posterior Sampling (DAPS) strategy \cite{zhang2025improving}. Unlike stepwise guidance approaches (e.g., DPS) that repeatedly inject observation corrections during sampling and may introduce artifacts or distupte the learned prior distribution \cite{schmidt2025generative,li2024learning}, DAPS avoids such instability by decoupling constraint enforcement from the diffusion trajectory and it enables the exploration of a larger solution space.

As illustrated in Figure \ref{fig3}(b), this method decouples the reverse diffusion step from the measurement consistency step, allowing for flexible and accurate guidance.The sampling process iterates from noise $\tau=1$ to clean data $\tau=0$. At each integration step $\tau$, the update consists of three distinct phases: Prior Estimation (Reverse Diddusion), Observation Guidance (Langevin Dynamics), and Noise Annealing.

First, we estimate the clean data $\hat{\mathbf{x}}_{0}$ from the current noisy state $\mathbf{x}_{\tau}$ using our pre-trained Prior Model. Since our model $\mathbf{s}_\phi$ is trained on local spatiotemporal cubes (Section \ref{subsec3_2}), we employ a sliding window inference strategy to ensure global consistency while preserving local temporal dependencies. For every time step $i$ in the sequence, we extract the local window $\mathbf{x}_{\tau}[i-k : i+k]$. This window is fed into the network $\mathbf{s}_\theta$ to predict the score. Crucially, to aggregate these predictions into a coherent global field, we only extract the score corresponding to the center frame (index $k+1$) of the output:

$$\nabla_{\mathbf{x}_{\tau}[i]} \log p(\mathbf{x}_{\tau}) \approx \mathbf{s}_\theta \Big( \mathbf{x}_{\tau}[i-k : i+k], \tau \Big) [k+1]$$

Using this synthesized global score, we apply a standard solver (e.g., Euler ODE solver) to predict the clean data estimate, denoted as $\hat{\mathbf{x}}_{0}(\mathbf{x}_{\tau})$. This represents the model's "best guess" of the AOD field based solely on learned prior knowledge, without yet considering the specific observations $\mathbf{y}$.

To incorporate the observational data $\mathbf{y}$, we construct an approximation of the posterior $p(\mathbf{x}_0 \mid \mathbf{x}_\tau, \mathbf{y})$. Following the DAPS formulation, we perform Langevin Dynamics initialized at the prior estimate $\hat{\mathbf{x}}_{0}(\mathbf{x}_{\tau})$. This process iteratively refines the estimate to satisfy the likelihood $p(\mathbf{y} \mid \mathbf{x})$ while remaining within the high-probability region of the prior. To approximate the intractable conditional distribution $p(\mathbf{x}_0 \mid \mathbf{x}_\tau)$, DAPS employs a Gaussian approximation centered at $\hat{\mathbf{x}}_{0}(\mathbf{x}_{\tau})$. Leveraging this, we define the guidance objective function $\mathcal{J}$ over the clean data estimate $\mathbf{x}_0$ to support multiple heterogeneous observation sources:

$$\mathcal{J}(\mathbf{x}_{0}) = \underbrace{ \frac{1}{2\sigma_{prior}^2} \| \mathbf{x}_{0} - \hat{\mathbf{x}}_{0}(\mathbf{x}_{\tau}) \|_2^2 }_{\text{Prior Consistency}} + \sum_{m=1}^{M} \lambda_m \underbrace{ \| \mathcal{A}_m(\mathbf{x}_{0}) - \mathbf{y}_m \|_2^2 }_{\text{Observation Fidelity}}$$

where $\mathbf{x}_{0}$ is the iterative variable in the Langevin loop, $\sigma_{prior}$ controls the adherence to the prior estimate. The second term aggregates constraints from $M$ observation sources, where $\mathcal{A}_m$ and $\lambda_m$ represent the forward operator and weight coefficient for the $m$-th observation source, respectively.

The estimate $\mathbf{x}_0$ is iteratively updated for $N$ steps using the Langevin Dynamics equation:

$$\mathbf{x}_{0}^{(j+1)} = \mathbf{x}_{0}^{(j)} - \eta \nabla_{\mathbf{x}_{0}} \mathcal{J}(\mathbf{x}_{0}^{(j)}) + \sqrt{2\eta}\boldsymbol{\epsilon}_j, \quad \boldsymbol{\epsilon}_j \sim \mathcal{N}(\mathbf{0}, \mathbf{I})$$

where $\eta$ is the step size. After $J$ iterations, the final sample $\mathbf{x}_{0}^{(J)}$ is treated as the observation-guided estimate, which we denote as $\mathbf{x}_{0}(\mathbf{x}_{\tau}, \mathbf{y})$.
The refined sample $\mathbf{x}_{0}(\mathbf{x}_{\tau}, \mathbf{y})$ represents the clean data at $\tau=0$. To proceed to the next diffusion time step $\tau - \Delta \tau$, we must decouple the consecutive samples as per the DAPS strategy. We reinject Gaussian noise into $\mathbf{x}_{0}(\mathbf{x}_{\tau}, \mathbf{y})$ scaled to the target noise variance $\sigma_{\tau - \Delta \tau}^2$:

$$\mathbf{x}_{\tau - \Delta \tau} = \hat{\mathbf{x}}_{0}(\mathbf{x}_{\tau}, \mathbf{y}) + \sigma_{\tau - \Delta \tau} \boldsymbol{\epsilon}, \quad \boldsymbol{\epsilon} \sim \mathcal{N}(\mathbf{0}, \mathbf{I})$$

This "annealing" step ensures the trajectory remains consistent with the diffusion schedule, allowing the process to iteratively refine the result from pure noise to a complete, high-fidelity AOD field.

\section{Result}\label{sec4}

\subsection{Evaluation of Learned Prior Distribution}\label{sec4_}

This section evaluates the effectiveness of the trained model (detailed in Section \ref{subsec3_2}) in learning the probability distribution of the AOD field, especially when the training data was incomplete. To verify the model's capacity to learn the prior distribution of the AOD field, we generate samples unconditionally and quantify the consistency between the learned distribution and the true AOD field distribution by comparing their statistical characteristics.

This study constructs the following sample sets for the calculation of statistical characteristics:

\begin{itemize}
    \item Ground Truth Sample Sets ($\mathcal{D}_{\text{GT}}$): 
        \begin{itemize}
            \item Base Set ($\mathcal{D}_{\text{GT}}^{\text{Base}}$): Consists of the complete AOD sequence for the year 2024. This set serves as the benchmark distribution for subsequent evaluations.
            \item Contrast Set ($\mathcal{D}_{\text{GT}}^{\text{Ctr}}$): Consists of the complete AOD sequence for the year 2023. This set is used to measure the natural variation range of the true data distribution under limited sample size, providing a reference scale for comparing the subsequent generated samples.
        \end{itemize}
    \item Generated Sample Sets ($\mathcal{D}_{\text{Gen}}$):
        \begin{itemize}
            \item Uncorrupted Training Generation ($\mathcal{D}_{\text{Gen}}^{\text{Unc}}$): One year-long AOD sequence sample generated unconditionally by the model trained on the uncorrupted AOD dataset.
            \item Corrupted Training Generation ($\mathcal{D}_{\text{Gen}}^{\text{Corr}}$): One year-long AOD sequence sample generated unconditionally by the model trained on the corrupted AOD dataset.
        \end{itemize}
\end{itemize}

Specifically, if the statistical characteristics of $\mathcal{D}_{\text{Gen}}$ closely match those of $\mathcal{D}_{\text{GT}}^{\text{Base}}$ and the discrepancy does not exceed the natural variation observed between $\mathcal{D}_{\text{GT}}^{\text{Ctr}}$ and $\mathcal{D}_{\text{GT}}^{\text{Base}}$, it strongly indicates that the model has successfully learned the prior distribution of the AOD field.
We first conduct a quantitative assessment of the generated samples sets  using common metrics from computer vision generation tasks. The results are presented in Table \ref{table4_1}, where the comparison is anchored against the Ground Truth Base Set ($\mathcal{D}_{\text{GT}}^{\text{Base}}$). FID quantifies the similarity between the distribution of generated samples and the real data in a deep feature space. Both $\mathcal{D}_{\text{Gen}}^{\text{Unc}}$ and $\mathcal{D}_{\text{Gen}}^{\text{Corr}}$ achieve comparable FID scores, affirming that the models have learned a distribution close to the true AOD field. Precision measures the proportion of generated samples falling near the manifold of real data, reflecting fidelity. Recall measures the extent to which the real data manifold is covered by the generated samples, reflecting diversity. $\mathcal{D}_{\text{Gen}}^{\text{Corr}}$ exhibits slightly lower Precision and Recall compared to $\mathcal{D}_{\text{Gen}}^{\text{Unc}}$, suggesting a minor reduction in both the coverage of the generated distribution due to training on inherently incomplete data. LPIPS is a perceptual distance metric used here to measure the average pairwise dissimilarity within a sample set. The similarity among generated samples is on par with the similarity among real samples. It indicated the model has not exhibited mode collapse, where generated samples become overly similar.

\begin{table}[h]
\centering
\caption{Comparison of Sample Sets using Generative Metrics}
\label{table4_1}
\begin{tabular}{c c c c c}
\toprule
\textbf{Sample Set} & \textbf{FID ($\downarrow$)} & \textbf{Precision ($\uparrow$)} & \textbf{Recall ($\uparrow$)} & \textbf{LPIPS ($\uparrow$)} \\
\midrule
$\mathcal{D}_{\text{GT}}^{\text{Ctr}}$ & 32.52 & 0.97 & 0.98 & 0.3356 \\
$\mathcal{D}_{\text{Gen}}^{\text{Unc}}$ & 38.14 & 0.91 & 0.92 & 0.3188 \\
$\mathcal{D}_{\text{Gen}}^{\text{Corr}}$ & 44.34 & 0.82 & 0.86 & 0.3015 \\
\bottomrule
\end{tabular}
\end{table}

\begin{figure}[ht]
  \centering
  \includegraphics[width=0.9\textwidth]{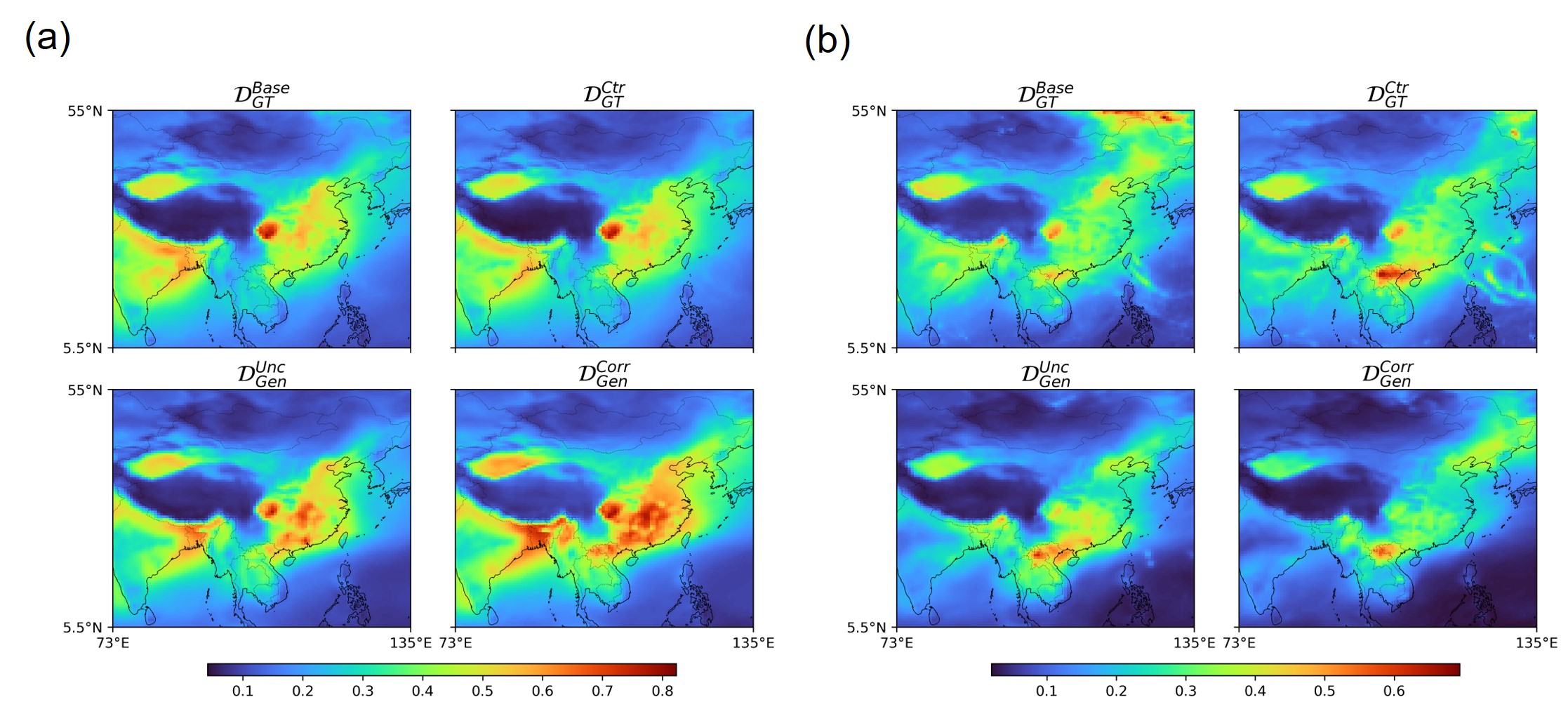}
  \caption{Comparison of (a) Spatial Mean and (b) Standard Deviation of  Ground Truth and Generated Sample Sets.}
  \label{fig4_1_1}
\end{figure}

Then We computed the spatial mean and standard deviation for each sample set to evaluate the generative model's ability to capture the macroscopic structure and spatial variability of the AOD field. Figure \ref{fig4_1_1}(a) presents the annual average spatial distribution of AOD across the four sample sets. The strong agreement between the generated and true data demonstrates the model's capacity to accurately reproduce the macroscopic spatial features of the AOD field. Crucial structures, such as the low-value AOD areas dictated by geographical features (e.g., the Tibetan Plateau) and the spatial extent of highly polluted regions, are successfully replicated in the generated mean fields ($\mathcal{D}_{\text{Gen}}^{\text{Unc}}$ and $\mathcal{D}_{\text{Gen}}^{\text{Corr}}$). Figure \ref{fig4_1_1}(b) illustrates the spatial standard deviation, which quantifies the spatiotemporal variability at each location. The spatial distribution of the standard deviation for the generated samples generally aligns well with the true data distribution, particularly in major high-variability areas like known pollution centers and regional boundaries. However, a closer examination reveals a subtle but important difference: the Ground Truth samples ($\mathcal{D}_{\text{GT}}^{\text{Base}}$ and $\mathcal{D}_{\text{GT}}^{\text{Ctr}}$) consistently exhibit more intense, localized high-value peaks in the standard deviation plots. In contrast, the corresponding Generated samples ($\mathcal{D}_{\text{Gen}}^{\text{Unc}}$ and $\mathcal{D}_{\text{Gen}}^{\text{Corr}}$) show a standard deviation pattern that is smoother and slightly less intense in these specific high-variability areas. This discrepancy suggests a potential smoothing effect inherent in the generation process, where the model slightly underestimates the variability associated with localized, extreme AOD events.

\begin{figure}[ht]
  \centering
  \includegraphics[width=0.9\textwidth]{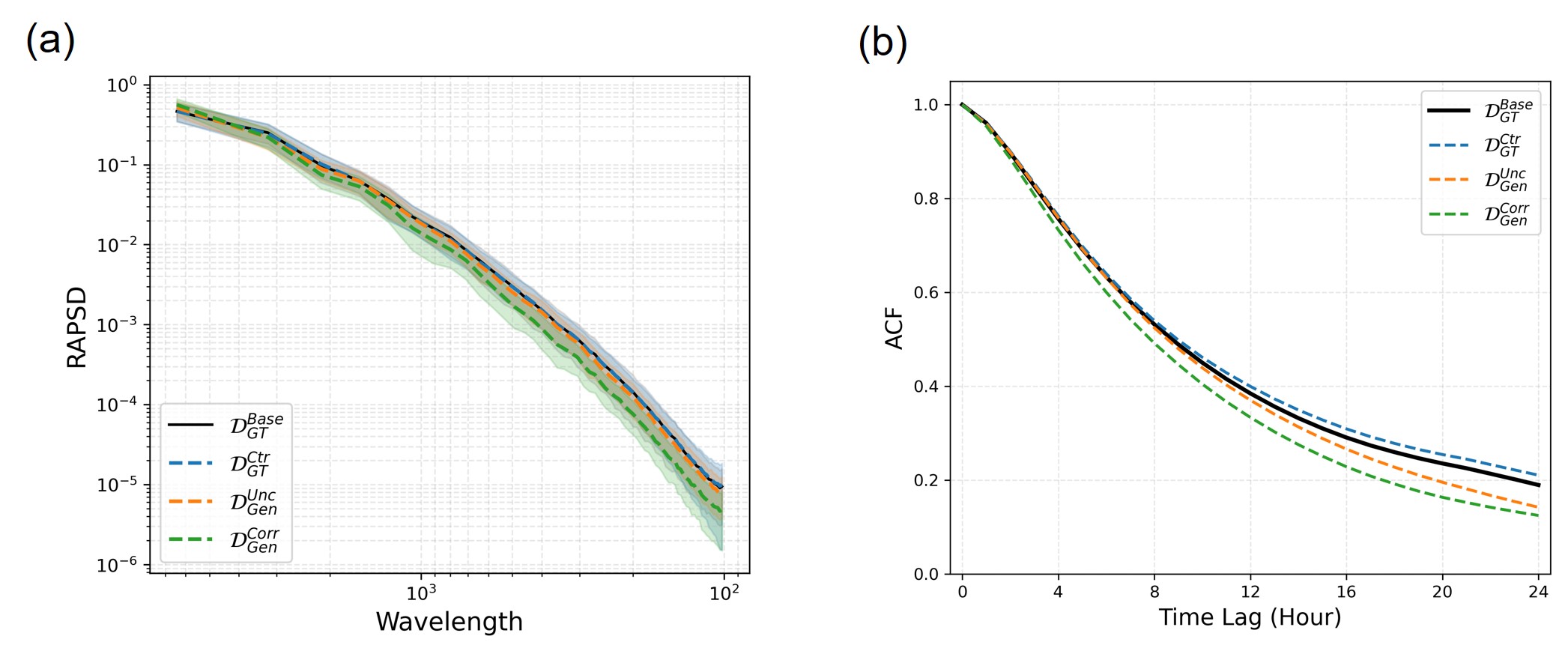}
  \caption{Comparison of (a) Rotational Average Power Spectral Density (RAPSD) and (b) Spatially Averaged Temporal Autocorrelation (ACF) across Ground Truth and Generated Sample Sets.}
  \label{fig4_1_2}
\end{figure}

We further examine the generated samples in the frequency and time domains by comparing the Rotational Average Power Spectral Density (RAPSD) and the Spatially. Averaged Temporal Autocorrelation Function (ACF). RAPSD provides a critical quantitative measure of how the variance (or "energy") is distributed across different spatial scales. Figure \ref{fig4_1_2}(a) demonstrates that the RAPSD curves for the generated samples ($\mathcal{D}_{\text{Gen}}$) closely align with the ground truth samples ($\mathcal{D}_{\text{GT}}$) across the entire spectrum. This high degree of spectral consistency confirms the model’s robust ability to accurately capture the distribution of energy from large-scale (low wavenumber) macroscopic structures down to small-scale (high wavenumber) fine details. This verifies that the model is generating textures and structures at all relevant spatial resolutions with high fidelity. To evaluate the model's fidelity in reproducing the temporal persistence characteristic of the AOD field, we computed the temporal autocorrelation function (ACF) for the sample sets. Figure \ref{fig4_1_2}(b) displays the ACF, which measures the correlation between the field at a given time and a later time (lag). The generated samples closely match the decay rate observed in the ground truth, particularly for short lags. This agreement indicates that the model successfully captures the physical memory of the AOD field, producing sequences that are statistically consistent and realistic over time.

Those results above show that the generated samples closely match the statistical characteristics of the true data. Even with incomplete training data, the model effectively learns the spatiotemporal probability distribution of AOD and can generate diverse, realistic samples through its powerful implicit modeling capability.

\subsection{Performance Evaluation of the Posterior Guidance Strategy}

This section evaluates the effectiveness of AODDiff on two core tasks critical for AOD reconstruction: Downscaling and Inpainting. The models are tested using a dataset covering the period from January to August 2025. We compare the performance of the decoupled annealing posterior sampling (DAPS) strategy (employed in our AODDiff) against the modified step-wise diffusion posterior sampling (DPS) strategy (employed in \cite{rozet2023score,schmidt2025generative}). Furthermore, we examine the impact of prior model selection by benchmarking reconstruction results generated from the prior trained on uncorrupted (clean) data ($\mathbf{s}_{\theta}^{\text{Unc}}$) against those from the prior trained on corrupted data ($\mathbf{s}_{\theta}^{\text{Corr}}$). Performance is quantified using two metrics:

\begin{itemize}

    \item \textbf{Normalized Root Mean Square Error (nRMSE)}: This metric measures the overall reconstruction fidelity by assessing the average error relative to the intrinsic variability of the data.
    $$\text{nRMSE} = \frac{1}{T} \sum_{t=1}^{T} \left( \frac{\sqrt{\frac{1}{N_{grid}} \sum_{i=1}^{N_{grid}} (\hat{x}_{t, i} - x_{t, i})^2}}{\sigma_{\mathbf{x}}} \right)$$
    where $\hat{x}_{t, i}$ and $x_{t, i}$ are the reconstructed and ground truth values at time $t$ and spatial location $i$, respectively, $N_{grid}$ is the total number of grid points, and $\sigma_{\mathbf{x}}$ is the standard deviation of the entire ground truth dataset.
    
    \item \textbf{Mean Energy Log-Ratio (MELR)}: This metric quantifies the model's ability to preserve the correct distribution of energy across various spatial scales (wavenumbers), thereby assessing spatial spectral fidelity. A lower MELR indicates that the generated samples accurately reproduce both the large-scale structures and the fine-scale texture of the AOD field. MELR is defined as the mean of the absolute logarithm ratio of the radial average power spectral densities (RAPSD):
    $$\text{MELR} = \frac{1}{T} \sum_{t=1}^{T} \left( \frac{1}{K} \sum_{k=1}^{K} \left| \log \left( \frac{\text{RAPSD}_{\text{Gen}}(t, k)}{\text{RAPSD}_{\text{GT}}(t, k)} \right) \right| \right)$$
    where $T$ is the total number of time steps in the evaluation period, $K$ is the total number of wavenumbers, and $\text{PSD}_{\text{Gen}}$ and $\text{PSD}_{\text{GT}}$ are the power spectral densities of the generated sample and the ground truth, respectively, at time $t$ and wavenumber $k$.
    
\end{itemize}

\begin{figure}[ht]
  \centering
  \includegraphics[width=0.9\textwidth]{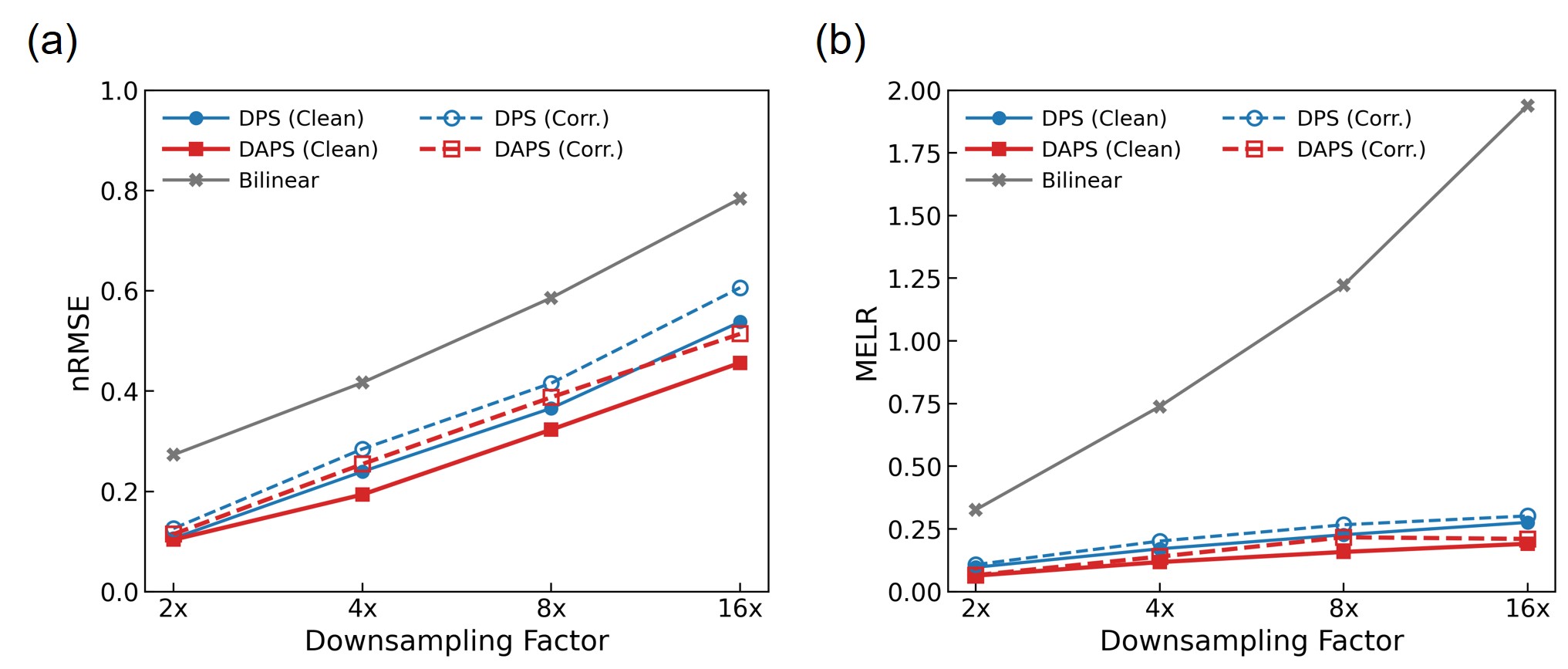}
  \caption{Comparison of (a) nRMSE and (b) MELR for different methods in the Downscaling task. }
  \label{fig4_2_1}
\end{figure}

\begin{figure}[ht]
  \centering
  \includegraphics[width=0.9\textwidth]{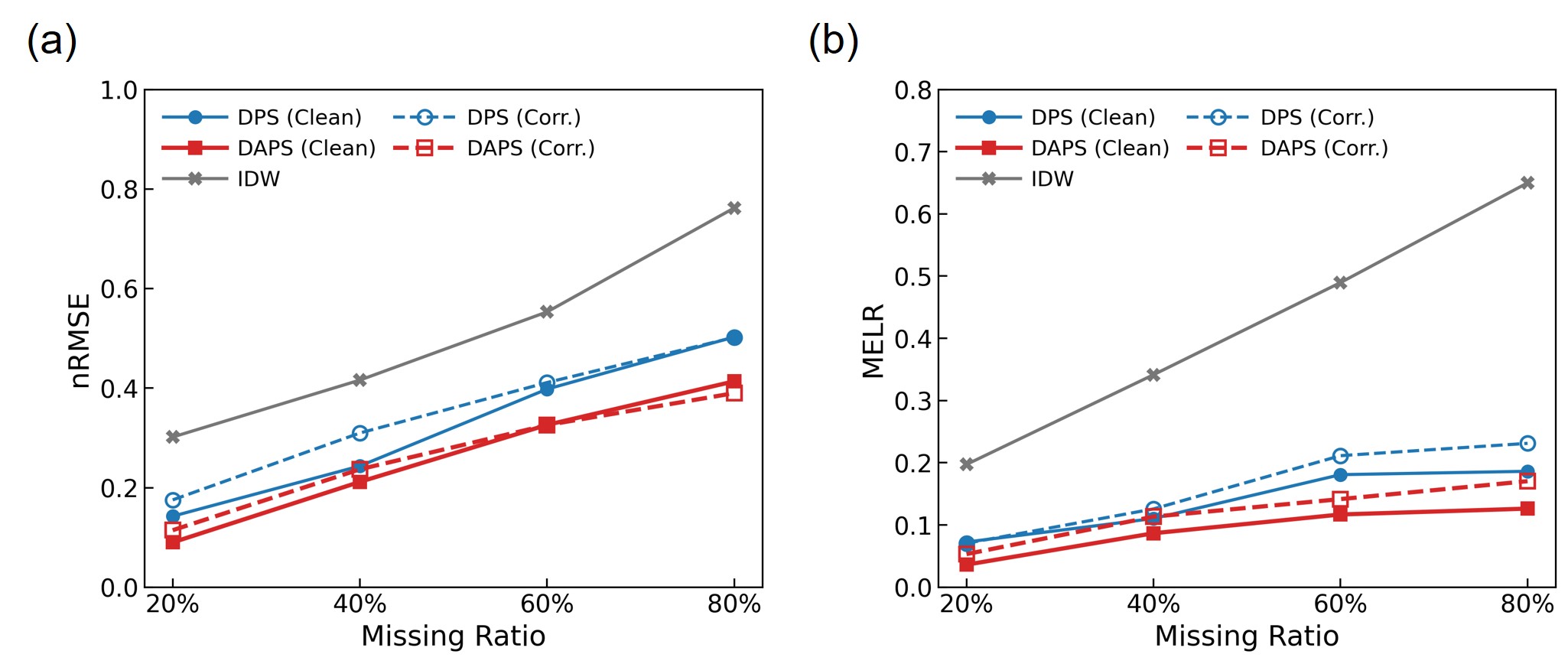}
  \caption{Comparison of (a) nRMSE and (b) MELR for different methods in the Inpainting task. }
  \label{fig4_2_2}
\end{figure}

Figures \ref{fig4_2_1} and \ref{fig4_2_2} present the quantitative performance of AODDiff across two core tasks: Downscaling and Inpainting, evaluating overall reconstruction fidelity (nRMSE) and spatial spectral preservation (MELR). A primary observation from both figures is that generative-based approaches significantly outperform traditional deterministic baselines. Notably, as the reconstruction difficulty intensifies (higher downsampling factors or increased missing ratios), the MELR of traditional methods exhibits a rapid surge. In contrast, generative models consistently maintain MELR at a much lower and stable level. This indicates that deterministic interpolation relies primarily on local smoothness assumptions and therefore fails to recover the fine-scale textures and high-frequency atmospheric structures inherent in AOD fields. By contrast, generative models leverage learned prior distributions to synthesize physically plausible high-frequency structures, thereby preserving high spectral fidelity even under severe data loss.

Regarding prior model, the results demonstrate that the prior trained on naturally corrupted data ($\mathbf{s}_{\theta}^{\text{Corr}}$) achieves performance highly comparable to the prior trained on uncorrupted data ($\mathbf{s}_{\theta}^{\text{Unc}}$). The absence of significant degradation confirms the effectiveness of the corruption-aware training strategy. Notably, in the Inpainting task (Figure \ref{fig4_2_2}a), when the missing ratio exceeds 60\%, the corrupted prior ($\mathbf{s}_{\theta}^{\text{Corr}}$) even exhibits a slightly lower nRMSE than the clean prior. This can be attributed to the fact that $\mathbf{s}_{\theta}^{\text{Corr}}$ is optimized directly on incomplete observations via corruption-aware training strategy. By learning from the specific noise and masking patterns present in real-world data, the model develops a stronger adaptive capability, allowing it to generalize better to high missing ratio scenarios. Furthermore, the DAPS strategy used in AODDiff consistently outperforms the modified DPS approach across all test cases. This suggests that the decoupled guidance mechanism is more effective for AOD field reconstruction than the stepwise injection used in DPS. While DPS often introduces smoothing artifacts or boundary instabilities due to its rigid guidance trajectory, DAPS facilitates a broader exploration of the solution space. It ensures that the generated AOD fields are not only numerically accurate but also structurally coherent, featuring sharp and realistic spectral characteristics.

\subsection{Case Study: Reconstruction Visualization and Uncertainty Analysis}

\begin{figure}[ht]
  \centering
  \includegraphics[width=0.95\textwidth]{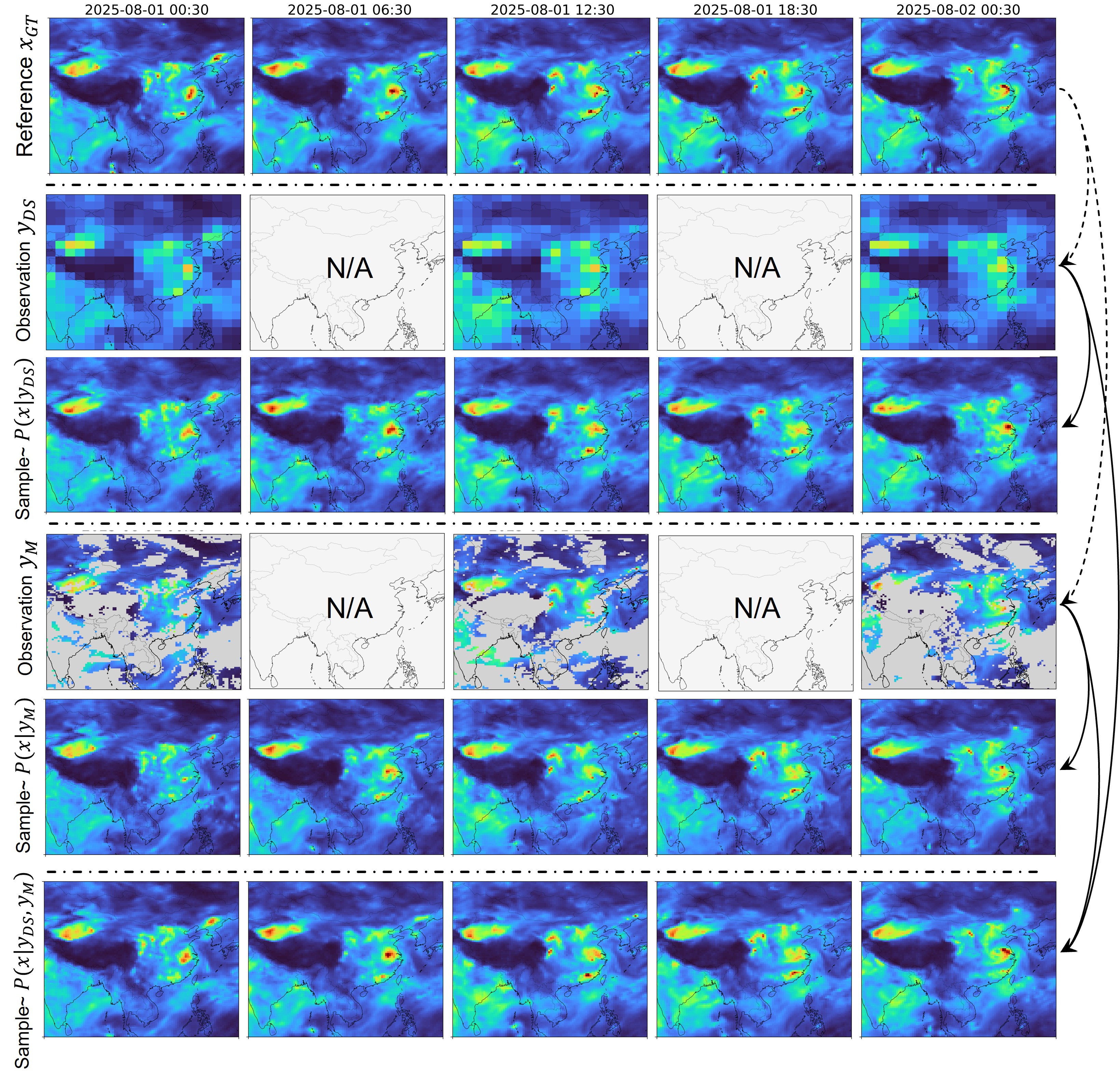}
  \caption{Visual Comparison of AODDiff Reconstruction under Different Observation Scenarios}
  \label{fig4_3_1}
\end{figure}

To provide a qualitative intuition of the AODDiff framework's performance and flexibility, we present a case study using a representative sequence from August 1st, 2025. As illustrated in Figure \ref{fig4_3_1}, the reconstruction result is visualized to demonstrate the model's behavior under various observational constraints. For all tasks in this section, we utilize the prior trained on naturally corrupted data ($\mathbf{s}_{\theta}^{\text{Corr}}$) combined with the DAPS guidance strategy to sample from the corresponding posterior distributions.

The first row shows the ground truth fields ($x_{GT}$) as a reference, where samples are visualized at 6-hour intervals to capture the spatiotemporal evolution of AOD distributions throughout the day. The second and fourth rows depict the simulated observations generated by the downscaling operator ($y_{DS}$, with a 4x spatial factor) and the masking operator ($y_M$, with an average missing ratio of approximately 45\%), respectively. To evaluate the model's ability to capture temporal dependencies, we applied temporal downsampling such that observations are provided only every four hours. This creates scenarios where no observational data is available at specific timestamps (e.g., 06:30 and 18:30, marked as N/A in the observation rows).

The third and fifth rows show samples drawn from the posterior distributions under the constraints of the downsampled observations ($Sample \sim P(x|y_{DS})$ and the masked observations ($Sample \sim P(x|y_M)$, respectively. The visualization results demonstrate that AODDiff successfully restores high-frequency details consistent with the reference $X_{GT}$ and effectively infills systematic missing regions. Even at timestamps without direct observations (marked as 'N/A'), the model leverages learned spatiotemporal dependencies to achieve accurate recovery, showcasing its generative capability in maintaining field continuity. The final row demonstrates the flexibility of AODDiff in simultaneously incorporating multiple heterogeneous observations. By sampling from the joint posterior distribution $P(x|y_{DS},y_M)$, the model effectively fuses the global structural guidance of the downsampled data ($y_{\text{DS}}$) with the local high-fidelity details from the unmasked portions of $y_M$. And this multi-source integration leads to significant improvements in reconstruction quality, exhibiting sharper spatial details and smoother temporal transitions compared to single-constraint results.

\begin{figure}[ht]
  \centering
  \includegraphics[width=0.95\textwidth]{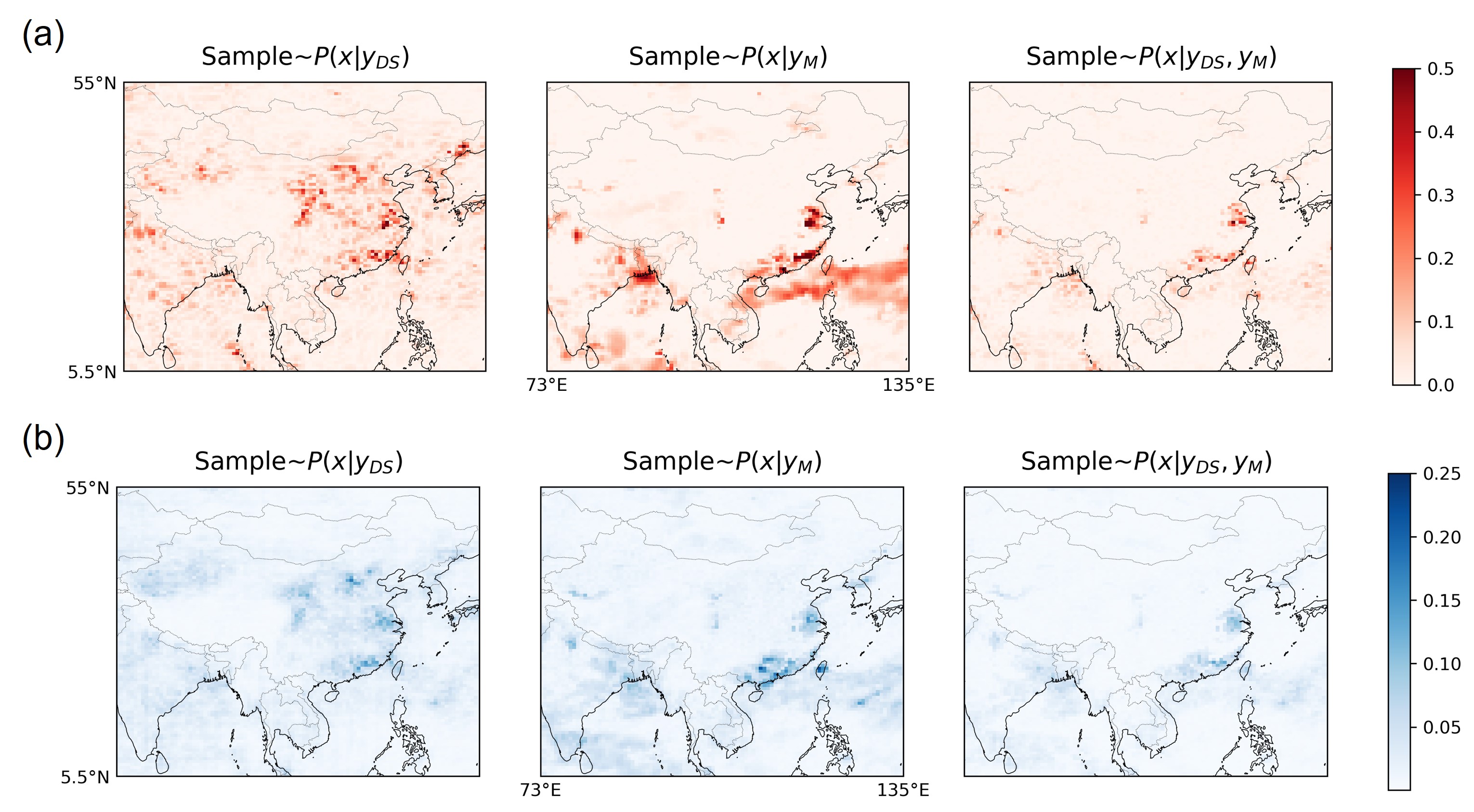}
  \caption{Spatial distribution of reconstruction error and predictive uncertainty at 12:30. (a): Pixel-wise absolute error maps. (b): Uncertainty maps (Standard Deviation) derived from multiple posterior samples. Columns represent the Downscaling task, Inpainting task, and Joint task, respectively.}
  \label{fig4_3_2}
\end{figure}

Figure \ref{fig4_3_2} illustrates the spatial distribution of reconstruction errors and predictive uncertainty at the 12:30 timestamp. The assessment was conducted by performing multiple posterior sampling operations to calculate the pixel-wise standard deviation, which serves as a metric for predictive uncertainty, while the reconstruction mean is compared against the ground truth to derive absolute error maps.

In the Downscaling task (left column), both the reconstruction error and predictive uncertainty are uniformly distributed across the spatial domain. This global consistency is expected, as the information loss associated with spatial downsampling is uniform, making high-frequency recovery equally challenging across the entire field. In contrast, the Inpainting task (middle column) exhibits a highly distinct error distribution, with significant concentrations of error and uncertainty localized within the masked regions. The deep blue areas in the uncertainty map highlight the ill-posed nature of inpainting in regions where no observational data is available to constrain the generative process. The advantage of the AODDiff is evident in the Joint task (right column), where the incorporation of dual observational constraints ($P(x|y_{DS}, y_M)$) significantly mitigates the aforementioned issues. The error maps show a marked reduction in intensity compared to single-task baselines, while the uncertainty maps demonstrate a substantial contraction in variance. Specifically, the low-frequency information provided by $y_{DS}$ effectively "anchors" the reconstruction in areas where $y_M$ is missing. This synergy between heterogeneous observation sources significantly reduces generative variance and confirms that the unified probabilistic framework effectively leverages diverse data modalities to enhance both reconstruction confidence and accuracy.

\section{Conclusion}

In this study, we proposed AODDiff, a diffusion-based Bayesian inference framework that reframes AOD reconstruction as a probabilistic generation task. By integrating a corruption-aware training strategy with decoupled nnealing posterior sampling strategy, the model successfully learns robust spatiotemporal priors directly from incomplete data and flexibly incorporates heterogeneous observational constraints without retraining. Extensive validation on MERRA-2 reanalysis data demonstrated the superior performance of AODDiff over traditional interpolation methods and standard diffusion guidance baselines. It overcomes the spectral smoothing inherent in traditional methods, preserving high-frequency spatial details even under severe data loss. And the framework supports the scalable integration of heterogeneous observation, leading to progressively improved reconstruction fidelity as more comprehensive observational information becomes available. Furthermore, the framework inherently quantifies predictive uncertainty by computing the standard deviation across multiple posterior samples, offering essential confidence metrics for risk-sensitive applications. Ultimately, AODDiff demonstrates significant potential for reconstructing spatiotemporally continuous AOD fields, offering a promising solution to the inherent incompleteness of current satellite retrievals to produce high-fidelity, gap-free aerosol records.

\paragraph{Acknowledgments}

This work was supported by the National Natural Science Foundation of China (Grant No. 32471866). The authors, except Hongqiang Fang, also thank the USTC Supercomputing Center for providing the necessary computational resources.

\bibliographystyle{unsrt} 

\bibliography{references}

\end{document}